\documentclass[journal]{IEEEtran}

\usepackage{mathtools}
\usepackage{bbm}
\usepackage{amsmath} 
\usepackage{bm}
\usepackage{graphicx}
\usepackage{hyperref}
\usepackage{todonotes}
\usepackage{multirow}
\usepackage[normalem]{ulem}
\usepackage{amsfonts}

\usepackage{graphicx}

\usepackage[noadjust]{cite}

\usepackage[acronym]{glossaries}

\newacronym{TDConv}{\textit{Time-Distributed Convolutions}}{\textit{Time-Distributed Convolutions}}
\newacronym{TDMaxp}{\textit{Time-Distributed Max Pooling}}{\textit{Time-Distributed Max Pooling}}
\newacronym{TDUps}{\textit{Time-Distributed Upsampling}}{\textit{Time-Distributed Upsampling}}
\newacronym{Conv2D}{\textit{2D Convolutions}}{\textit{2D Convolutions}}
\newacronym{Bidirconvlstm}{\textit{Bidirectional ConvLSTM}}{\textit{Bidirectional C-LSTM}}
\newacronym{concat}{\textit{Concatenation}}{\textit{Concatenation}}

\newacronym{CLSTM}{\textit{C-LSTM}}{\textit{Convolutional Long Short-Term Memory}}
\newacronym{FCN}{\textit{FCN}}{\textit{Fully-Convolutional Networks}}
\newacronym{LSTM}{\textit{LSTM}}{\textit{Long Short-Term Memory}}
\newacronym{CNN}{\textit{CNN}}{\textit{Convolutional Neural Networks}}
\newacronym{Sensor3D}{\textit{Sensor3D}}{\textit{Sensor3D}}

\ifCLASSINFOpdf
\else
\fi
\begin{document}
	%
	\title{Deep Sequential Segmentation of Organs in Volumetric Medical Scans}
	%
	%
	%
	%
	%
	
	\author{Alexey A.~Novikov,
		David~Major,
		Maria~Wimmer,
		Dimitrios~Lenis,
		Katja~B\"{u}hler
		\thanks{}
		\thanks{A. A. Novikov, D. Major, M. Wimmer, D. Lenis and K. B\"{u}hler are with the VRVis Zentrum f\"{u}r Virtual Reality und Visualisierung Forschungs-GmnH, 1220 Vienna, Austria, e-mail: (novikov@vrvis.at, major@vrvis.at, mwimmer@vrvis.at, lenis@vrvis.at, buehler@vrvis.at). VRVis is funded by BMVIT, BMDW, Styria, SFG and Vienna Business Agency in the scope of COMET - Competence Centers for Excellent Technologies (854174) which is managed by FFG. Thanks go to our project partner AGFA HealthCare for valuable input. Copyright (c) 2018 IEEE. Personal use of this material is permitted. However, permission to use this material for any other purposes must be obtained from the IEEE by sending a request to pubs-permissions@ieee.org.}}
	
	\maketitle              
	
	\begin{abstract}
		
	Segmentation in 3D scans is playing an increasingly important role in current clinical practice supporting diagnosis, tissue quantification, or treatment planning. The current 3D approaches based on \gls{CNN} usually suffer from at least three main issues caused predominantly by implementation  constraints - first, they require resizing the volume to the lower-resolutional reference dimensions, second, the capacity of such approaches is very limited due to memory restrictions, and third, all slices of volumes have to be available at any given training or testing time. We address these problems by a U-Net-like~\cite{Ronneberger2015} architecture consisting of bidirectional \gls{CLSTM}~\cite{Shi2015} and convolutional, pooling, upsampling and concatenation layers enclosed into time-distributed wrappers. Our network can either process the full volumes in a sequential manner, or segment slabs of slices on demand. We demonstrate performance of our architecture on vertebrae and liver segmentation tasks in 3D CT scans.  
		
	\end{abstract}
	\section{Introduction}
	
	Accurate segmentation of anatomical structures in volumetric medical scans is of high interest in current clinical practice as it plays an important role in many tasks involved in computer-aided diagnosis, image-guided interventions, radiotherapy and radiology. In particular, quantitative diagnostics requires accurate boundaries of anatomical organs.  
	
	Computed tomography (CT) is currently among the most used 3D imaging modalities. Despite its inability of differentiating organs with similar intensities it is widely used for diagnosis of diseases in organs. Manual segmentation in CT can be a very tedious task. Therefore, automated methods with minor or no human interaction at all, are preferable.
	
	
	Automated segmentation with deep learning methods in medical images has popularized widely in the recent years, mainly due to the success of applying \acrfull{FCN} in natural images~\cite{Long2015} and consequently in the biomedical imaging~\cite{Ronneberger2015}. Since then various modifications of \acrshort{FCN}s have been proposed for segmentation of different anatomical organs and imaging modalities. 
	
	
	3D scans are generally represented as stacks of 2D images. Running a segmentation algorithm on the 2D slices directly with merging results afterwards ignores spatial inter-slice correlations, therefore hybrid 2D/3D and direct 3D approaches gained popularity. Most of these methods are built upon 2D~\cite{Ronneberger2015} and 3D~\cite{Cicek2016} U-Net architectures. Lu et al.~\cite{Lu2017} proposed to locate and segment the liver via convolutional neural networks and graph cuts. Dou et al. \cite{Dou2016} presented a 3D \acrshort{FCN} which boosts liver segmentation accuracy by deep supervision layers. Yang et al. \cite{Yang2017} used adversarial training in order to gain in performance for the 3D U-Net segmentation of the liver in CT scans. Sekuboyina et al.~\cite{Sekuboyina2017} proposed a pipeline approach for both localization and segmentation of the spine in CT. Here the vertebrae segmentation is performed in a blockwise manner to overcome memory limitations as well as obtain a fine-grained result. A similar blockwise approach in combination with a multi-scale two-way \gls{CNN} was introduced by Korez et al.~\cite{Korez2017}.

	Other noteworthy works using variants of 2D and 3D U-Nets consider applications in cardiac MR image segmentation~\cite{Baumgartner2018, Yu2017}, pancreas in 3D CT~\cite{Zhou2017, Heinrich2017} and prostate in 3D MR~\cite{Yu2017Prostate, Liu2018}. A variety of papers have contributed to the multiple tasks in brain imaging such as segmentation of cerebrospinal fluid, gray and white matter~\cite{Chen2018}, brain tumour~\cite{Dong2017, Shen2017}, multiple sclerosis lesion~\cite{Brosch2016} and glioblastoma~\cite{Yi2016}.  
	
	In order to overcome memory limitations modern \gls{CNN}-based methods are usually preceded by downsampling of the input scans. This might result in a deformation of organs in the image, causing information loss.
	
	Consequently, hybrid 2D/3D methods that process volumetric data in a slice-wise fashion followed by a 3D processing step, gained importance. For instance, Li et al.~\cite{Li2017} applied a slice-wise densely connected variant of the 2D U-Net architecture for liver segmentation first, and refined the result by the auto-context algorithm in 3D. For the same task, Christ et al.~\cite{Christ2016} applied slice-wise 2D U-Nets to obtain a rough segmentation first, and then tuned the result in a second step with Conditional Random Fields. Relying only on intra-slice data is insufficient for proper leveraging spatial information. In order to address this issue, the above-mentioned methods applied computationally expensive 3D classical image processing refinement strategies in addition to the 2D \acrshort{CNN}-based approach. 
	
	Hybrid approaches combining \acrshort{FCN} with recurrent networks such as \gls{LSTM}~\cite{Hochreiter1997} and more recently proposed \acrshort{CLSTM}~\cite{Shi2015} are effective for processing sequential data in general. Hence, the recurrent networks have recently been introduced to the biomedical imaging context. A method proposed by Poudel et al.~\cite{Poudel2017} uses a U-Net variant to get an estimate of the 2D slice-wise segmentation, which is subsequently refined by the so-called gated recurrent unit~\cite{Cho2014}, a simplified version of the \gls{LSTM}.
	
	Bates et al.~\cite{Bates2017} evaluated several architectures involving both \acrshort{CLSTM} and standard convolutional layers. In the \textit{deep} configuration, several bidirectional \acrshort{CLSTM} units were stacked in the U-shaped architecture in which the outputs of the forward and backward \acrshort{LSTM} passes were concatenated. In the \textit{shallow} configuration a shared copy of the CNN was applied to each slice of the 3D scans separately and then the result was passed to the stacked \acrshort{CLSTM} units to produce the segmentation volume. For the purpose of designing a multi-scale architecture, Chen et al.~\cite{Chen2016} used a variant of 2D U-Net to extract features for all slices first, and then processed them with bidirectional \acrshort{CLSTM} units in order to exploit 3D context.

	Though the described approaches address some issues of the deep learning based 3D segmentation algorithms such as voxel size anisotropy and intensive computations due to 3D convolutions, they still do not take into account the two following issues. First, they require that all volumes have the same fixed input reference dimensions, and, second, all slices of the volumes have to be available in order to extract 3D context at both training and testing time. The former scenario is not always applicable usually due to large variations of the number of slices in the volumes across even the same dataset, and the latter one could force reducing network capacity due to memory and timing restrictions what could potentially lead to lower accuracies.
	
	To overcome these problems we propose to integrate bidirectional \acrshort{CLSTM}s into a U-Net-like architecture in order to extract the 3D context of slices in a sequential manner. In this way, the network is able to learn the inter-slice correlations based on the slabs of the volume. The downsampling of the input is not required anymore as only a fraction of the volume is processed at any given time. Training of this network is therefore not demanding memory-wise which is another known limitation of the current modern networks. This fully integrated sequential approach can be particularly useful for real-time applications as it enables segmentation already during data acquisition or while loading the data as both are generally performed slice-by-slice.
	
	Furthermore, we show the invariance of our method to field-of-view and orientation by evaluating on two CT datasets depicting two different organs, namely liver and vertebrae. 
	
	For the sake of simplicity of the further explanation in the following we refer to this architecture as \acrshort{Sensor3D} (acronym for "sequential segmentation of organs in 3D").

	\begin{figure*}[th!]
		\label{fig:overview}
		\includegraphics[width=\linewidth]{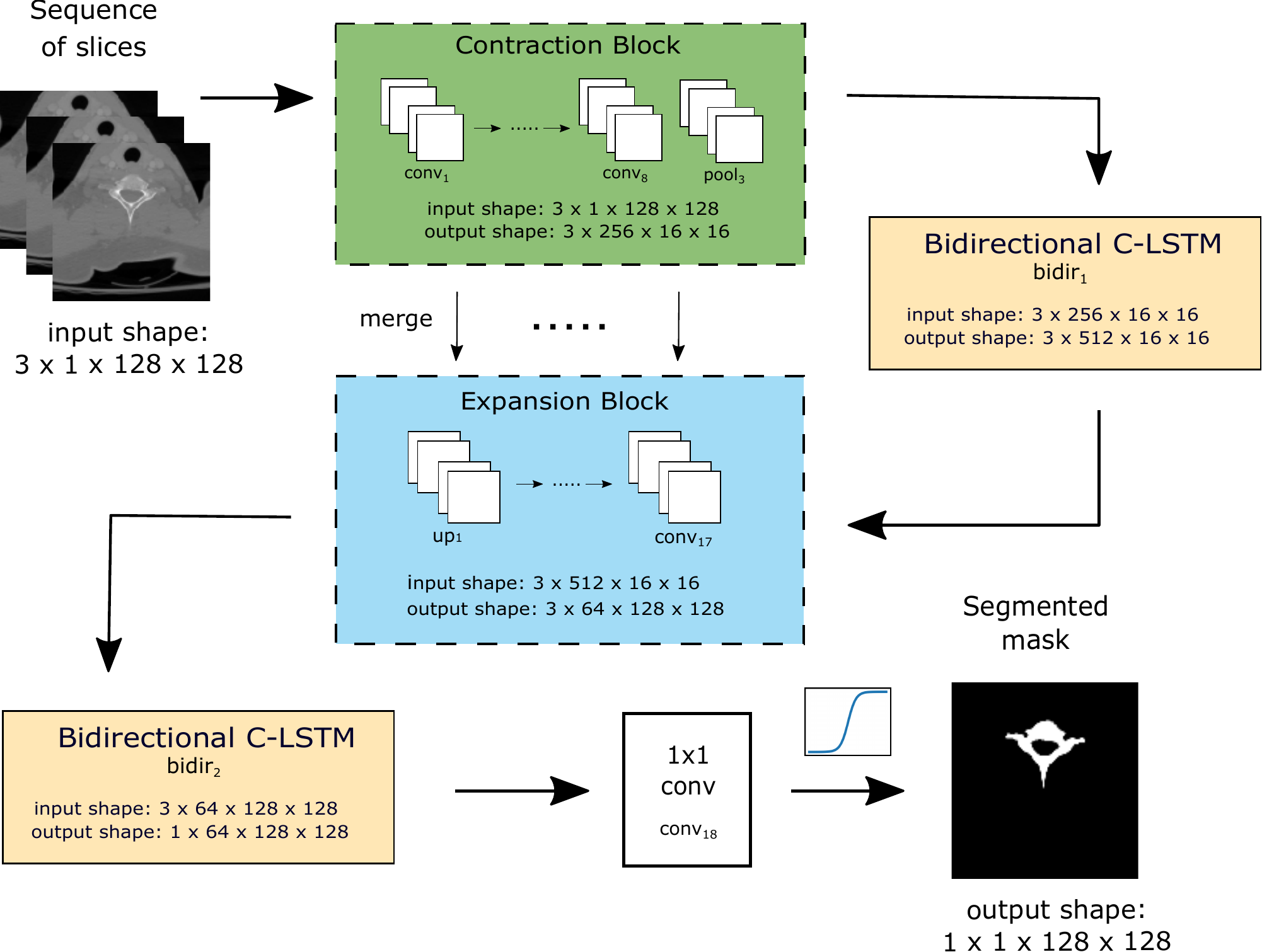}
		\caption{Overview of the proposed \acrshort{Sensor3D} architecture for a sample of three vertebrae slices and $128\times128$ imaging resolution used during training. Contraction and Expansion blocks are enclosed into time-distributed wrappers. Dashed merge connection corresponds to concatenations between layers of contraction and expansion blocks. The layer names in the network blocks correspond to entries in Table~\ref{table:architecture_details}.}
	\end{figure*}

	\section{Methodology}

	\subsection{General setup}
	\label{sec:main_notations}
	
	Let $\mathcal{I}=\{I_1, ... , I_n \}$ be a set of $n \in \mathbb{N}$ volumetric scans, where each $I_i$, $i=1,...,n$ consists of voxels $\bm{x} = (x_1, x_2, x_3) \in \mathbb{R}^3$ with intensities $I_i(\bm{x}) \in \mathcal{J} \subset \mathbb{R}$. More specifically, each scan $I_i$ is therefore a set of $m_i \in \mathbb{N}$ slices $J_k^i, k=1,...,m_i$ within the organ area where $J_k^i(\bm{y}) \in \mathcal{J}$ correspond to intensities at the pixels with positions $\bm{y} = (y_1, y_2) \in \mathbb{R}^2$ at the $k$-th slice of the scan $I_i \in \mathcal{I}$. 
	
	For each slice $J_k^i \in I_i$, a set of ground truth masks $M_{k}^i \coloneqq (M^i_{k,\,l})_{l=1} ^m$ is available, where $l$ corresponds to semantic class labels $\mathcal{L} = \{l_1, ..., l_m \}$ and $M^i_{k,\cdot} \in \mathcal{M}$ to the space of all 2D binary masks of the same size as the slices $J^i_k$.
	
	
	To enforce reproducibility of the input flow shapes, we build a new training dataset in the following way. 	
	
	The spatial context $C_k^i$ of the slice $J_k^i$ is defined as a set containing the slice $J_k^i$ and its $(\mathbf{o}-1)/2$ neighbouring slices above and below, selected equidistantly with a rounded stepsize $\mathbf{d}$ and the pre-defined length of the input sequence $\mathbf{o}$. Rounding to the more distant slice is performed if the precise step is not possible. Training set $\mathcal{I}^\prime$ is defined then as follows:
	\begin{equation}
	\mathcal{I}^\prime = \{C_k^i \mid i=1,..,n, k=1,..,m_i\}
	\end{equation} 
	
	For training and evaluation purposes, the dataset $\mathcal{I}^\prime$ is split into non-overlapping sets, namely  $\mathbf{I}^\prime_{\, \text{TRAIN}}$ and $\mathbf{I}^\prime_{\, \text{TEST}}$. During training, the network is consecutively passed through with minibatches $\mathcal{K} \in \mathcal{N}$, where $\mathcal{N}$ is a complete partition of the set $\mathbf{I}^\prime_{\, \text{TRAIN}}$.
	
	For each spatial context $C^i_k \in \mathcal{I}^\prime$, i.e. $C^i_k = \{J^i_p, ... , J^i_q\}$ for some $1\le p, q \le m_i$, the multi-class output of the network is calculated: understanding the network as a function 
	\begin{equation}
	\label{eq:network_function}
	\mathcal{N} : \mathcal{I}^\prime \rightarrow \mathcal{M},  
	\end{equation}
	$\mathcal{N}(C^i_k)$ derives for each pixel  $\bm{y} \in J^i_t$ its semantic class $l \in \mathcal{L}$ in a single step with some probability, where $J^i_t$ corresponds to the middle element of the spatial context $C^i_k$. In order to estimate and maximize this probability, we define a loss function
	\begin{equation}
	\Lambda :  \mathcal{I}^\prime \times \mathcal{M} \rightarrow \mathbf{R}
	\end{equation}
	that estimates the deviation (error) of the network outcome from the desired ground truth. Using the formal notations derived in our work~\cite{Novikov2018} we define the loss function in the following way.  
	
	
	For a distance function $d:\mathcal{I}^\prime  \times \mathcal{M}  \rightarrow \mathbf{R}$, weighting coefficients $r_{\mathcal{K}, l}$ and a spatial context $C^i_k \in \mathcal{K}$ the loss function is 
	\begin{equation}
	\label{eq:loss}
	\Lambda(C^i_k, M^i_k) \coloneqq -\sum_{l \in \mathcal{L}} r_{\mathcal{K}, l}^{-1} \,\, d(C^i_k,  M^i_k ) 
	\end{equation}
	over the set $\mathcal{K}$ and the complete partition. 
	
	
	
	The distance function $d_l^{\,dice}$ for the Dice coefficient for a spatial context $C^i_k$, a feature channel $l$, ground-truth mask $M^i_k$ and sigmoid activation function $p_l(\cdot)$ can then be defined as:
	\begin{equation}
	d_l ^{\,dice} ( C^i_k,   M^i_k) :=  2 \, \frac{ \sum_{\bm{x} \in I} \chi_{\pi_l(M^i_k)}(\bm{x}) \, p_l(\bm{x})}{ \sum_{\bm{x} \in I} \left(\chi_{\pi_l(M^i_k)}(\bm{x}) + p_l(\bm{x})\right)}
	\end{equation}
	where $\chi_{\pi_l(M^i_k)}(\bm{x})$ is a characteristic function, i.e., $\chi_{\pi_l(M^i_k)}(\bm{x}) = 1$ iff  $M^i_k$ is $1$ at position of pixel $\bm{x}$ and $0$ otherwise. The definition of the loss function in this equation would allow for using multiple classes, however, this is beyond the scope of this work.

	\subsection{Building the architecture}
	\label{sec:architecture}
	
	Following the above, a 3D volumetric scan $I_i$ can be interpreted as a \mbox{time-series} of 2D slices $\{J_1, ... , J_{m_i}\}$. Such series can then be processed using methods known for successful performance on sequential data. The time-distributed convolutions and recurrent networks in particular are a natural choice for such 3D scans. Time-distributed convolutions are typical convolutions passed to a time-distributed wrapper that allows application of any layer to every temporal frame (or slice) of the input independently. In the context of this work such temporal frames correspond to the elements of training sequences extracted from the volumes. In our architecture the wrapper was applied to all convolutional, pooling, upsampling and concatenation layers.
	
	In order to leverage spatio-temporal correlations of the order-preserving slices (that is elements of the $C^i_k$) and due to their sequential nature, we have combined the time-distributed layers and bidirectional \glspl{CLSTM} in an end-to-end trainable U-Net-like hybrid architecture. Main intuition for designing this architecture was that the features of the correlated slices should also be correlated. The \glspl{CLSTM} in our model are used to impose this correlation explicitly. To make training faster and to reduce the number of parameters, our \gls{CLSTM} blocks are based on the version of the \gls{LSTM} without connections from the cell to the gates (widely known as "peephole connections"). Motivation for using this variant was the research by Greff et al.~\cite{Greff2015} where it was shown that removing peephole connections in general does not hurt the overall performance.
		
	Fig.~\ref{fig:overview} shows the high-level overview of the proposed architecture on a sample sequence of vertebrae slices. Table~\ref{table:architecture_details} complements the figure with tensor shapes for each layer for a particular case when the length of input sequences $\mathbf{o}$ is equal to three. 
	
	As mentioned previously, the network takes an odd-lengthed spatial context $C^i_k$ as the input. This sequence is then passed to the contraction block (green in Fig.~\ref{fig:overview} and the corresponding layers from $conv_1$ to $pool_3$ in Table~\ref{table:architecture_details}). As all convolutional and max pooling layers are enclosed into a time-distributed wrapper, each element of the sequence is processed through the contraction block independently.  
	
	In order to capture spatio-temporal correlations between slices the features extracted for each element of the input sequence are passed into the \acrshort{CLSTM} block \cite{Shi2015} at the end of the contraction part (layer $bidir_1$ in Table~\ref{table:architecture_details}). In order to enable the network to learn spatio-temporal correlations of the slices in both directions, we used a bidirectional extension for the \acrshort{CLSTM} with the summation operator combining forward and backward outputs. This \acrshort{CLSTM} block aims at adding the explicit dependency of the low-dimensional high abstract features extracted for the elements of the sequence.   
	
	The sequence output of the bidirectional \acrshort{CLSTM} block is then passed to the expansion part (blue in Fig.~\ref{fig:overview} and the corresponding layers from $up_1$ to $conv_{17}$ in Table~\ref{table:architecture_details}). Similarly to the contraction part, each element of the sequence is processed independently via time-distributed convolutional as well as upsampling layers. After every upsampling layer, the features are concatenated with the corresponding features from the contraction part. When the spatial resolution of the features reaches the desired output sizes, the sequence is passed to another bidirectional \acrshort{CLSTM} block (layer $bidir_2$ in Table~\ref{table:architecture_details}). The sequence is processed in both directions and the outputs are combined by summation. At this stage this block contributes towards two goals: adding explicit dependency for the high-dimensional high-abstract features and converting the incoming sequence into a single-channelled output.  The resulting features are then passed to the (1,1) convolution layer in order to map each feature vector to the desired number of classes (in the scope of this work the number of classes is equal to one). The output of the last convolutional layer (layer $conv_{18}$ in Table~\ref{table:architecture_details}) is mapped into [0,1] range via the sigmoid activation which is applied to each pixel independently. This results in the segmentation of the middle element of the spatial context $C^i_k$.

	
	\begin{table*}[th!]
		\centering
		\caption{Detailed information on the proposed architecture with filters and shapes for input and output tensors for the case when the length of input sequences is $\mathbf{o}=3$ and in-plane imaging resolution is $128\times128$ }
		\label{table:architecture_details}
		\begin{tabular}{ccccc}
			\hline
			Layer Name & Layer Type & Input Shape & Filters & Output Shape \\ \hline
			$conv_1$ & \acrshort{TDConv} & $3\times1 \times 128 \times 128$ & $3 \times 64 \times 3 \times 3$ & $3 \times 64 \times 128 \times 128$ \\
			$conv_2$ & \acrshort{TDConv} & $3 \times 64 \times 128 \times 128$ & $3 \times 64 \times 3 \times 3$ & $3 \times 64 \times 128 \times 128$ \\
			$pool_1$ & \acrshort{TDMaxp} & $3 \times 64 \times 128 \times 128$ & $3 \times 2 \times 2$ & $3 \times 64 \times 64 \times 64$ \\
			$conv_4$ & \acrshort{TDConv} & $3 \times 64 \times 64 \times 64$ & $3 \times 128 \times 3 \times 3$ & $3 \times 128 \times 64 \times 64$ \\
			$conv_5$ & \acrshort{TDConv} & $3 \times 128 \times 64 \times 64$ & $3 \times 128 \times 3 \times 3$ & 3 $\times 128 \times 64 \times 64$ \\
			$pool_2$ & \acrshort{TDMaxp} & $3 \times 128 \times 64 \times 64$ & $3 \times 2 \times 2$ & $3 \times 128 \times 32 \times 32$ \\
			$conv_7$ & \acrshort{TDConv} & $3 \times 256 \times 32 \times 32$ & $3 \times 256 \times 3 \times 3$ & $3 \times 256 \times 32 \times 32$ \\
			$conv_8$ & \acrshort{TDConv} & $3 \times 256 \times 32 \times 32$ & $3 \times 256 \times 3 \times 3$ & $3 \times 256 \times 32 \times 32$ \\
			$pool_3$ & \acrshort{TDMaxp} & $3 \times 256 \times 32 \times 32$ & $3 \times 2 \times 2$ & $3 \times 256 \times 16 \times 16$ \\
			\hline \hline 
			$bidir_1$ & \textit{Bidirectional C-LSTM} & $3 \times 256 \times 16 \times 16$ & $512 \times 3 \times 3$ & $3 \times 512 \times 16 \times 16$ \\
			\hline \hline 
			$up_1$ & \acrshort{TDUps} & $3 \times 512 \times 16 \times 16$ & $3 \times 2 \times 2$ & $3 \times 512 \times 32 \times 32$ \\
			$concat_1$ & \acrshort{concat} ($conv_8$, $up_1$) &  &  & $3 \times 768 \times 32 \times 32$ \\
			$conv_{11}$ & \acrshort{TDConv} & $3 \times 768 \times 32 \times 32$ & $256 \times 3 \times 3$ & $3 \times 256 \times 32 \times 32$ \\
			$conv_{12}$ & \acrshort{TDConv} & $3 \times 256 \times 32 \times 32$ & $256 \times 3 \times 3$ & $3 \times 256 \times 32 \times 32$ \\
			$up_2$ & \acrshort{TDUps} & $3 \times 256 \times 32 \times 32$ & $3 \times 2 \times 2$ & $3 \times 256 \times 64 \times 64$ \\
			$concat_2$ & \acrshort{concat} ($conv_5$, $up_2$) &  &  & $3 \times 384 \times 64 \times 64$ \\
			$conv_{14}$ & \acrshort{TDConv} & $3 \times 384 \times 64 \times 64$ & $128 \times 3 \times 3$ & $3 \times 128 \times 64 \times 64$ \\
			$conv_{15}$ & \acrshort{TDConv} & $3 \times 128 \times 64 \times 64$ & $128 \times 3 \times 3$ & $3 \times 128 \times 64 \times 64$ \\
			$up_3$ & \acrshort{TDUps} & $3 \times 128 \times 64 \times 64$ & $3 \times 2 \times 2$ & $3 \times 128 \times 128 \times 128$ \\
			$concat_3$ & \acrshort{concat} ($conv_2$, $up_3$) &  &  & $3 \times 192 \times 128 \times 128$ \\
			$conv_{17}$ & \acrshort{TDConv} & $3 \times 192 \times 128 \times 128$ & $64 \times 3 \times 3$ & $3 \times 64 \times 128 \times 128$ \\
			\hline \hline 
			$bidir_2$ & \textit{Bidirectional C-LSTM} & $3 \times 64 \times 128 \times 128$ & $64 \times 3 \times 3$ & $1 \times 64 \times 128 \times 128$ \\
			\hline \hline 
			$conv_{18}$ & \acrshort{Conv2D} & $1 \times 64 \times 128 \times 128$ & $1 \times 1 \times 1$ & $1 \times 1 \times 128 \times 128$ \\ \hline
		\end{tabular}%
	\end{table*}

	\section{Experimental Setup}
	
	To evaluate the performance and generalizability of our architecture we trained an tested it for 3D segmentation of two different anatomical organs: liver and vertebrae in CT scans. Liver segmentation is often a required step in the diagnosis of hepatic diseases while the segmentation of vertebrae is important for the identification of spine abnormalities, e.g. fractures, or image-guided spine intervention.
	
	\subsection{Training data and preparation}
	
	For \textit{liver} segmentation we used two related datasets: 3Dircadb-01 and 3Dircadb-02 \cite{Soler2010} combined together. The first consists of 20 3D CT scans with hepatic tumours in 75\% cases. The second one consists of two anonymized scans with hepatic focal nodular hyperplasia. The axial in-plane resolution varied between 0.56 and 0.961 $mm^2$ and the slice thickness varied between 1.0 and 4.0 $mm$. The consecutive elements within the training sequences were generated at distances $\mathbf{d}\in \{3,5,7,9\} \, mm$ within the liver area. These numbers were chosen based on the maximal slice thicknesses in the scans of the dataset. Unlike other existing liver datasets, 3Dircadb is more challenging due to the presence of multiple pathological cases with tumours both inside and close to the liver. The whole dataset with annotations of different organs is publicly available. Detailed per-scan information is available online~\cite{3Dircadb}.
	
	We used a normalization technique similar to the one proposed by Christ et al.~\cite{Christ2016} which we applied to each slice of the sequences independently. First, the raw slices were windowed to [-100, 400] to prevent including non-liver organs. Second, the contrast-limited adaptive histogram equalization was applied to the clipped slices. Third, the results were zero-centered by subtracting the slice-wise mean and then additionally normalized by scaling using the slice-wise standard deviation. 
	
	For \textit{vertebrae} segmentation we used the CSI 2014 challenge train set~\cite{Yao2012}. It comprises 10 CT scans covering the entire lumbar and thoracic spine as well as full vertebrae segmentation masks for each scan. The axial in-plane resolution varies between 0.3125 and 0.3616 $mm^2$. The slice thickness is 1~$mm$. The consecutive elements within the training sequences were generated at the distances of $1 \,mm$ within the vertebrae area. 
	
	
	In this work we focused on learning the 3D spatial context in a direct neighbourhood to the slices of interest only, thus in all evaluations we used sequences of three slices $\mathbf{o}=3$. The design of the suggested architecture would allow for using larger sequences, however, this is beyond the scope of this work. 
	
	In order to prevent over-fitting for both liver and vertebrae segmentation tasks we made sure that every scan was first assigned either to the training or the testing set and only then converted into sequences. In this way, we ensured independence of the sets allowing us to estimate the generalizability of the algorithm.
	
	All slices and their corresponding masks in the training set were downsampled to $128\times128$ in-plane imaging resolution. In order to compute the performance scores resulting masks were upsampled to the original $512 \times 512$ imaging resolution during testing.

	
	
	

	\subsection{Training strategies}
	
	We trained the networks in an end-to-end manner over the loss shown by Eq.~\ref{eq:loss} using the Adam \cite{Kingma2014} optimization algorithm with a fixed initial rate of $5 \times 10^{-5}$ and the standard values of $\beta_1=0.9$ and $\beta_2=0.999$. Early stopping with the patience parameter equal to 100 epochs was used in all evaluations. Therefore, number of epochs varied between training runs.
	
	The learning rate was chosen empirically based on the preliminary evaluations on smaller training sets. Higher learning rates caused the network training to diverge whereas lower ones slowed it down significantly.
	 
	We used zero-padding in convolutional layers and \acrshort{CLSTM} in the Sensor3D and its variants in all evaluation runs. Therefore, output channels of the layer had the same dimensions as the input.
	
	Initialization with a random orthogonal matrix was used for the weights at the recurrent connections of the \acrshort{CLSTM} \cite{Saxe2013}. Glorot uniform \cite{Glorot2010} was utilized as an initialization for the weights at all other connections at the \acrshort{CLSTM} and at all convolutional layers.
	
	As activation function at all convolutional layers we employed exponential linear units~\cite{Clevert2015}. For the \acrshort{CLSTM} layers we used the widely used setup of hyperbolic tangent functions in all cases except the recurrent connections where the hard sigmoid was applied. 
	
	\subsection{Implementation Details}
	
	All experiments were performed using Keras with TensorFlow backend in Python. The backend was used for automatic differentiation and optimization during training. 
	
	Downsampling of the ground-truth masks and upsampling of the segmentation masks were performed using the transform module of the \emph{scikit-image} library.

	\subsection{Performance metrics}
	
	To evaluate the architectures and compare with state-of-the-art approaches, we used the Dice ($D$) similarity coefficient and volume overlap error ($VOE$), defined as follows. 
	
	Given an image $I$ and the feature channel $l$, let $\pi_l(M_I)$ be a set of foreground pixels in the channel $l$ of the ground-truth mask $M_I$ and $P_l(I)$ be the set of pixels where the model is certain that they do not belong to the background, i.e.,
	\begin{equation}
	P_l(I):=\left\{\bm{x}: \bm{x} \in I \, \land \, | \, p_l(\bm{x}) - 1 \, | < \epsilon \right\}
	\label{eq:pixel_set}
	\end{equation}
	where $\epsilon=0.25$ is an empirically chosen threshold value and $p_l(\bm{x})$ is the approximated probability of the pixel $\bm{x}$ belonging to the foreground.
	
	The coefficients $D$ and $VOE$ might then be computed in the following way:
	\begin{equation}
	\label{eq:dice}
	D ( I,  M_I) := 2 \, \frac{| \, P_l(I) \cap  \pi_l(M_I)  \, |}{|P_l(I)| + |\, \pi_l(M_I)|}
	\end{equation}
	\begin{equation}
	\label{eq:voe}
	VOE( I,  M_I) = \frac{2 \, (1-D( I,  M_I))}{2-D( I,  M_I)}
	\end{equation}
	
	\section{Results and Discussion}
	
	\begin{table*}[t!]
		\centering
		\caption{Detailed segmentation results of two-fold evaluations of liver segmentation task for different inter-slice distances}
		\label{table:two_fold_results}
		\begin{tabular}{|c|c|c|c|c|c|c|c|c|}
			\hline
			& \multicolumn{4}{c|}{Fold 1} & \multicolumn{4}{c|}{Fold 2} \\ \hline
			& \multicolumn{2}{c|}{Organ Area} & \multicolumn{2}{c|}{Full Volume} & \multicolumn{2}{c|}{Organ Area} & \multicolumn{2}{c|}{Full Volume} \\ \hline
			Step size & $D$ (\%) & $VOE$ (\%) & $D$ (\%) & $VOE$ (\%) & $D$ (\%) & $VOE$ (\%) & $D$ (\%) & $VOE$ (\%) \\ \hline 
			3 $mm$ & 94.8 & 9.8  & 92.8 & 13.4 & 95.1 & 9.4 & 93.7 & 11.8 \\ \hline 
			5 $mm$ & 95.5 & 8.6 & 94.1 & 11.1 & 96.1 & 7.5 & 95.6 & 8.4 \\ \hline
			7 $mm$ & 95.3 & 8.9 & 94.3  & 10.8 & 96.4 & 6.9 & 96.2 & 7.3 \\ \hline
			9 $mm$ & 95.5 & 8.6 & 94.6 & 10.2 & 96.4 & 6.9 & 96.2 & 7.3 \\ \hline
		\end{tabular}
	\end{table*}
	
	\begin{table*}[th!]
		\centering
		\caption{The significance difference analysis of segmentation results using Wilcoxon signed-rank test for Dice scores on the test set for the liver segmentation task. The p-values are given for fold 1 and fold 2 (separated by "slash" sign)}
		\begin{tabular}{|c|c|c|c|c|}
			\hline
			& 3 $mm$ & 5 $mm$ & 7 $mm$ & 9 $mm$ \\ \hline
			3 $mm$ & $\infty$ & \bm{$<0.01$} / \bm{$<0.01$} & \bm{$<0.01$} / \bm{$<0.01$} & \bm{$<0.01$} / \bm{$<0.01$} \\ \hline
			5 $mm$ & \bm{$<0.01$} / \bm{$<0.01$} &$\infty$  & 0.17 / 0.08  &  0.13 / 0.07 \\ \hline
			7 $mm$ & \bm{$<0.01$} / \bm{$<0.01$}  & 0.17 / 0.08  & $\infty$ & 0.85 / 0.3 \\ \hline
			9 $mm$ & \bm{$<0.01$} / \bm{$<0.01$} & 0.13 / 0.07  & 0.85 / 0.3 & $\infty$ \\ \hline
		\end{tabular}
		\label{table:significance}
	\end{table*}

	\begin{table*}[t!]
		\centering
		\caption{Detailed segmentation results of two-fold evaluations for architectures with different number of features in the convolutional layers and \acrshort{CLSTM} for the liver segmentation task}
		\label{table:capacity_results}
		\begin{tabular}{|c|c|c|c|c|c|c|c|c|}
			\hline
			& \multicolumn{4}{c|}{Fold 1} & \multicolumn{4}{c|}{Fold 2} \\ \hline
			& \multicolumn{2}{c|}{Organ Area} & \multicolumn{2}{c|}{Full Volume} & \multicolumn{2}{c|}{Organ Area} & \multicolumn{2}{c|}{Full Volume} \\ \hline
			 & $D$ (\%) & $VOE$ (\%) & $D$ (\%) & $VOE$ (\%) & $D$ (\%) & $VOE$ (\%) & $D$ (\%) & $VOE$ (\%) \\ \hline 
			Original configuration & 95.3 & 8.9 & 94.3  & 10.8 & 96.4 & 6.9 & 96.2 & 7.3 \\ \hline
			2$\times$ smaller & 95.3 & 8.9  & 93.9 & 11.5 & 96.2 & 7.3  & 95.9 & 7.9\\ \hline 
			4$\times$ smaller & 94.5 & 10.4 & 93.6 & 12.0 & 95.6 & 8.4 & 95.4 & 8.8 \\ \hline
			8$\times$ smaller & 94.3 & 10.8 & 92.6 & 13.8 & 94.6 & 10.2  & 94.3 & 10.8 \\ \hline
		\end{tabular}
	\end{table*}

	\begin{figure*}[t!]
		\centering
		\includegraphics[width=0.2\textwidth]{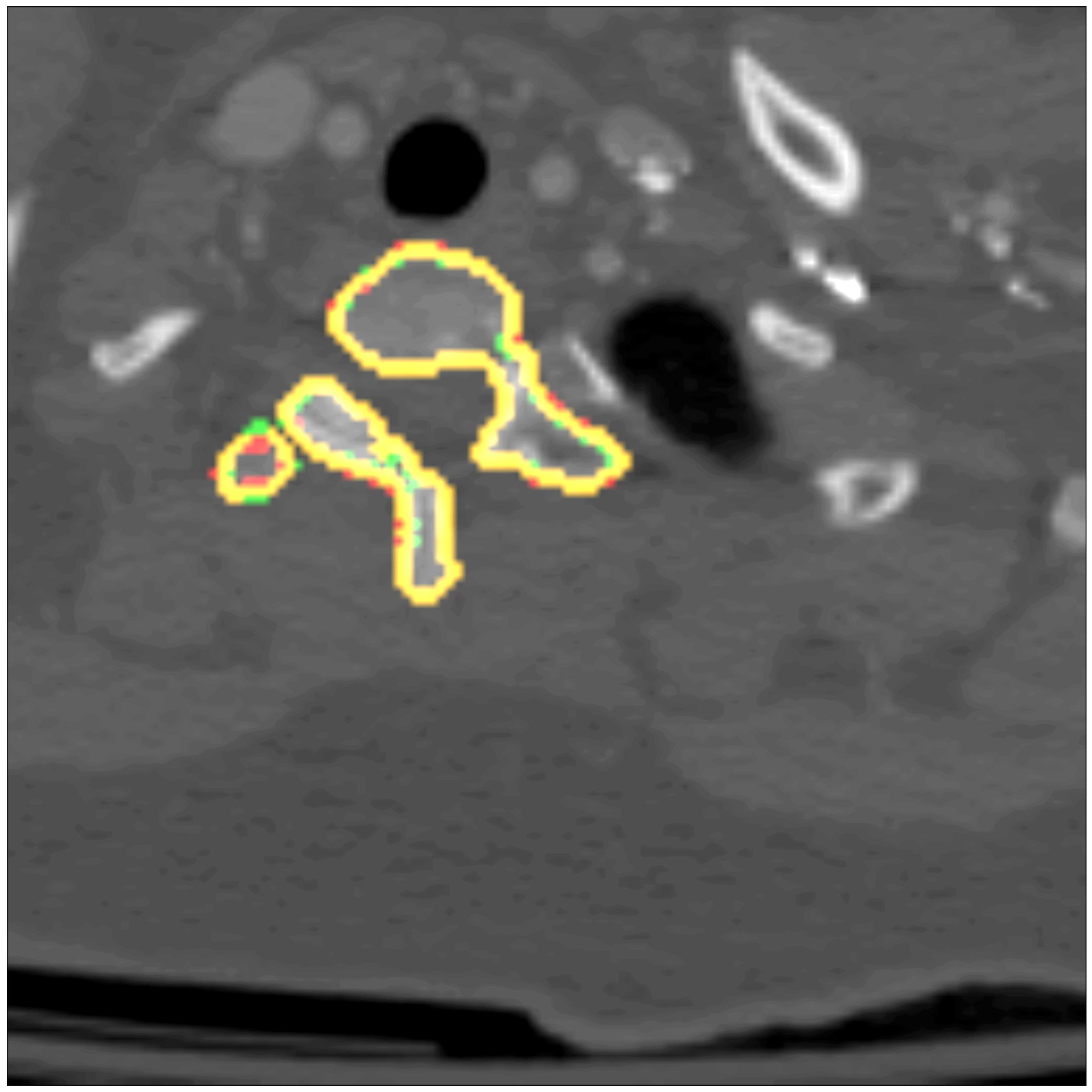}
		\includegraphics[width=0.2\textwidth]{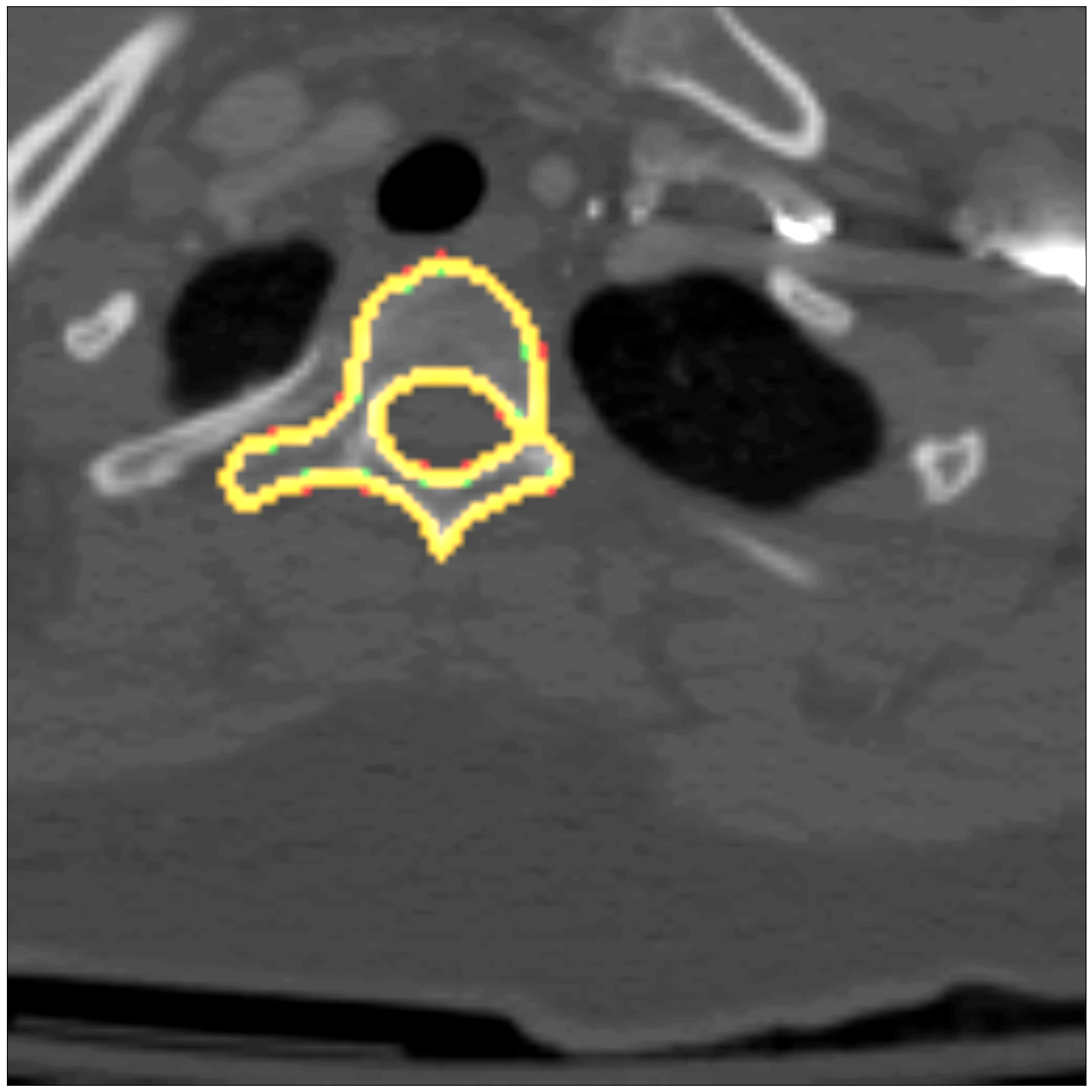}
		\includegraphics[width=0.2\textwidth]{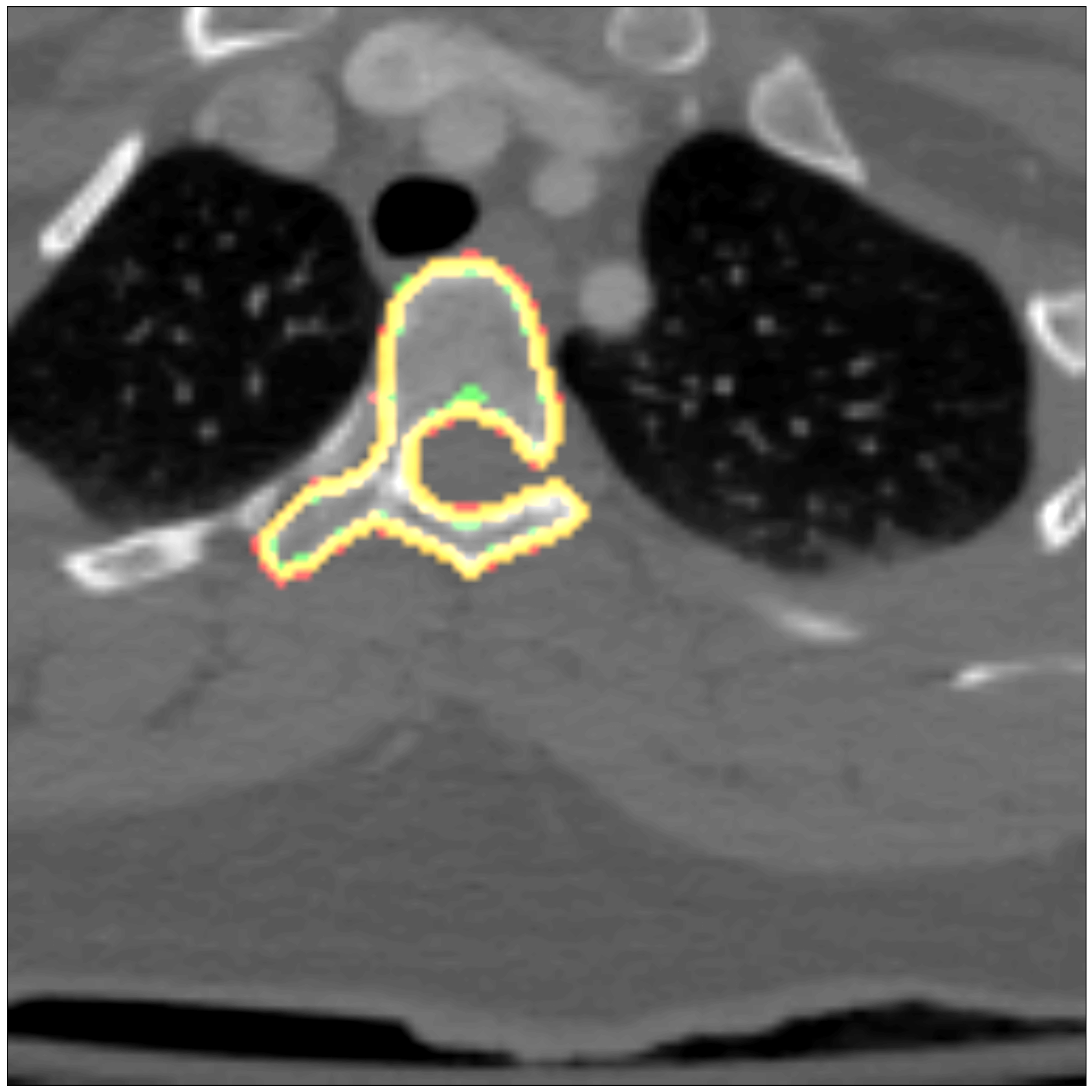}
		\includegraphics[width=0.2\textwidth]{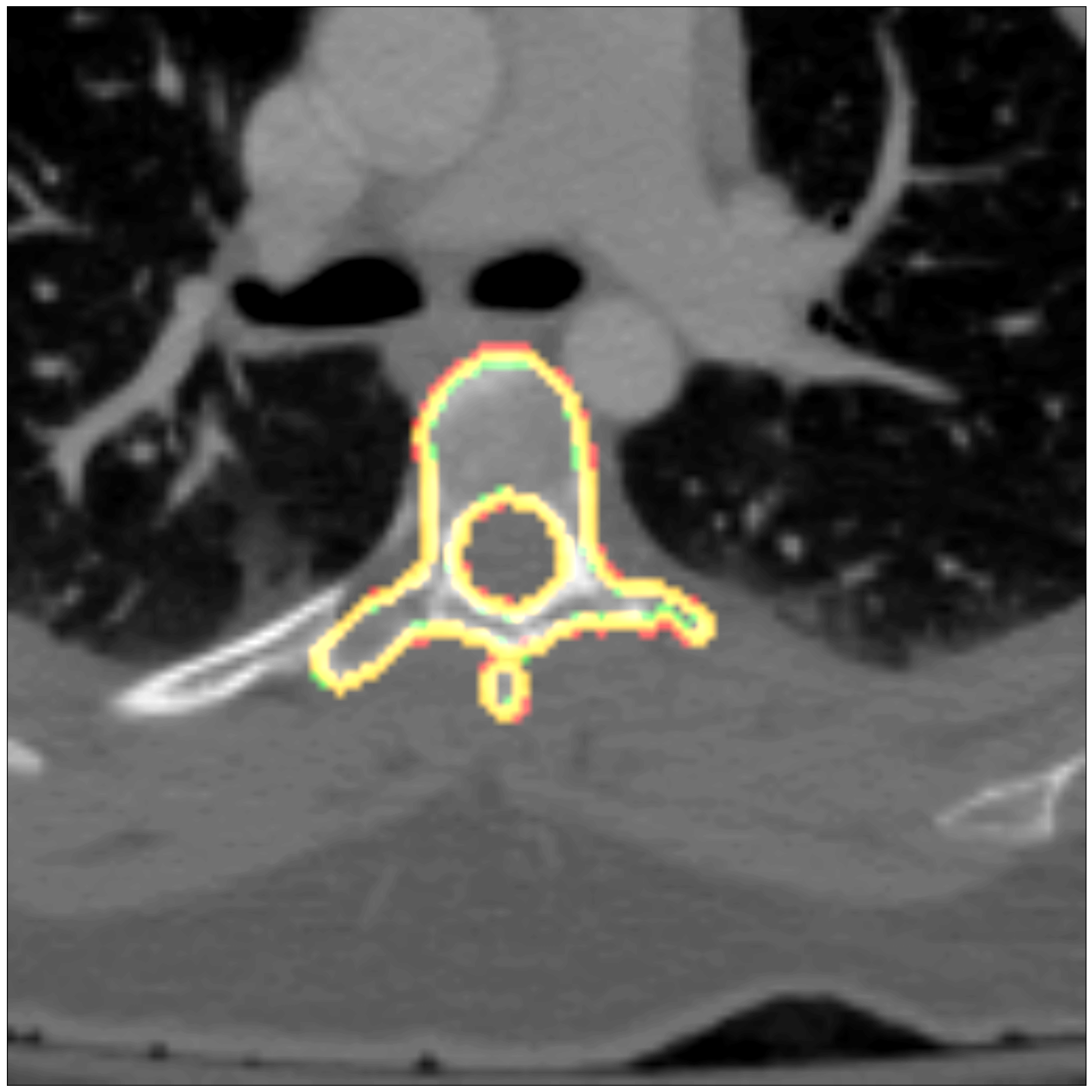}
		\includegraphics[width=0.2\textwidth]{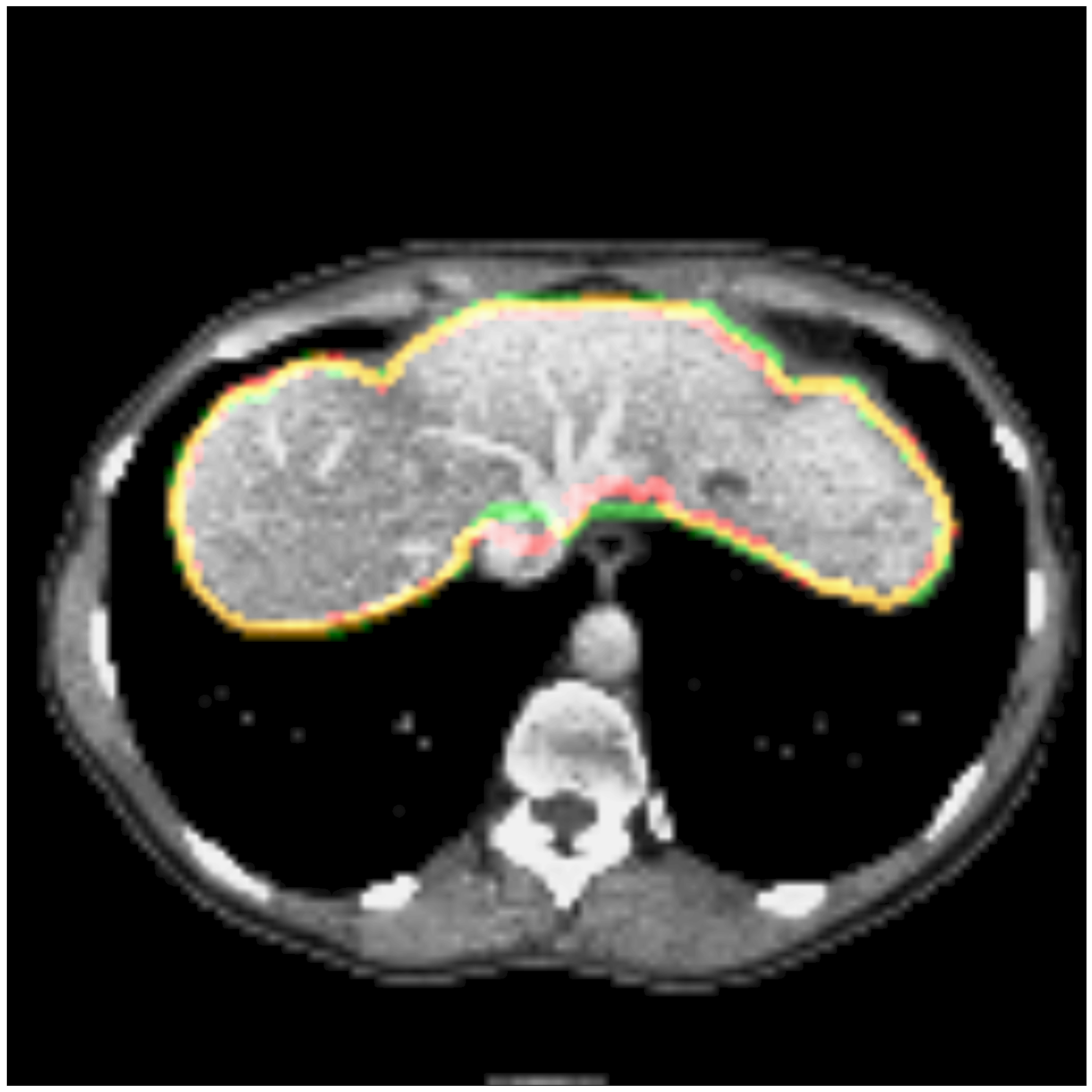}
		\includegraphics[width=0.2\textwidth]{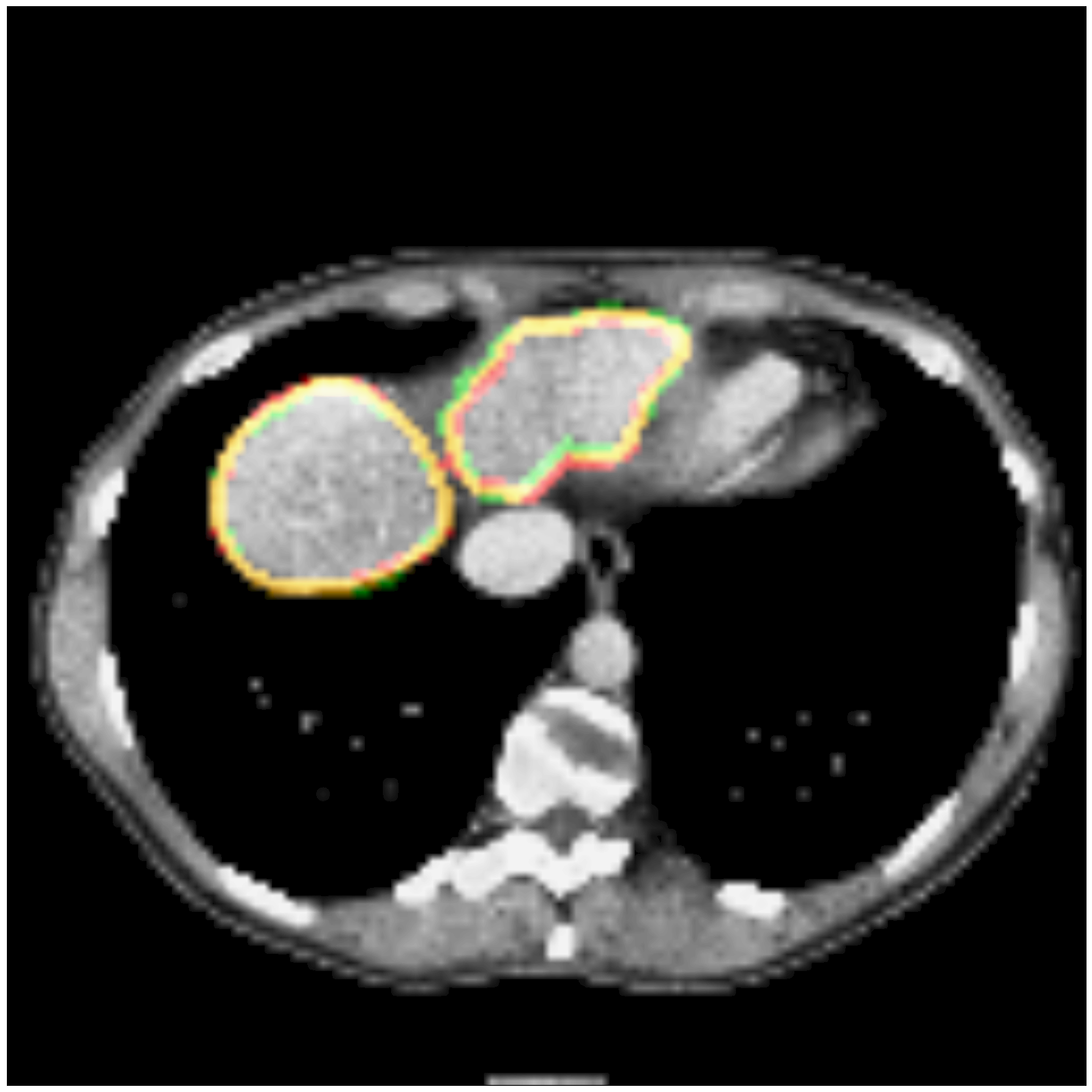}
		\includegraphics[width=0.2\textwidth]{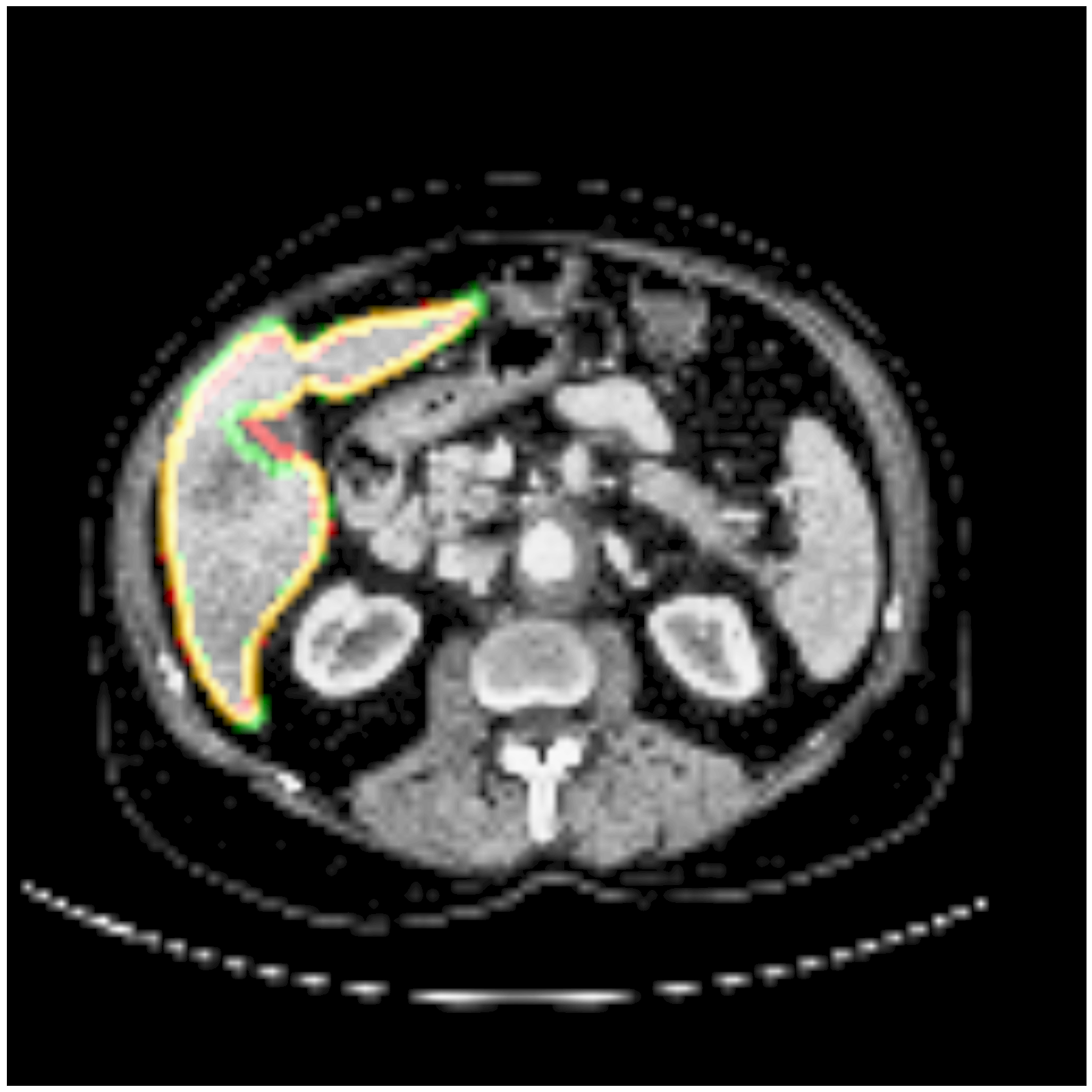}
		\includegraphics[width=0.2\textwidth]{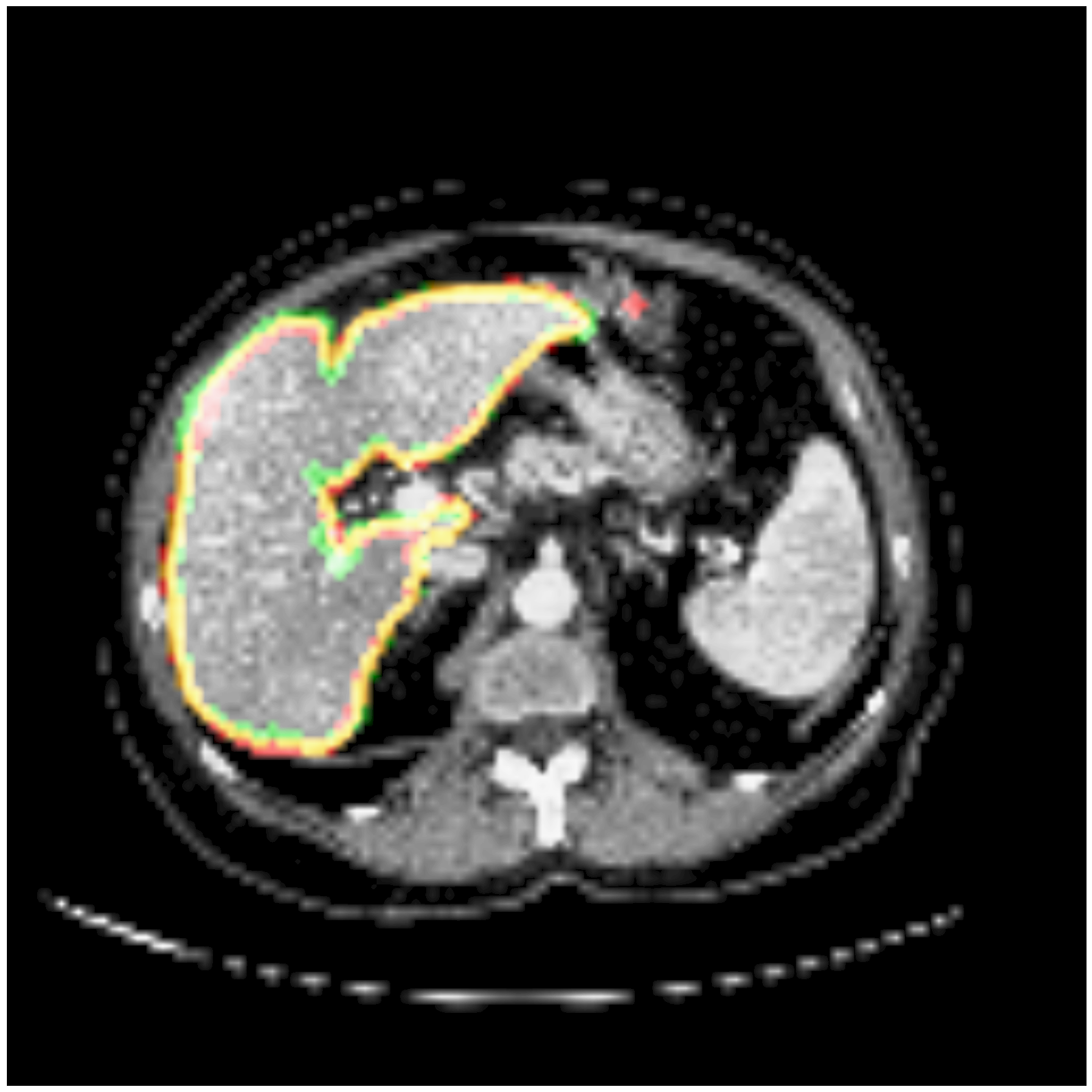}
		\caption{Several visual examples of segmentations results in different vertebrae (top) and liver (bottom) locations. The contour in red corresponds to the outline of the prediction, green to the ground-truth and yellow to the overlap of the outlines}
		\label{fig:segmentations}
	\end{figure*}

	

	\begin{table*}[ht]
		\caption{Our method compared with state-of-the-art methods on the liver segmentation on 3Dircadb (left) and vertebrae segmentation on CSI 2014 (right) datasets; "*" the score estimated using Eq.~\ref{eq:dice} or Eq.~\ref{eq:voe}; "**" the area of vertebrae available in ground-truth data}
		\label{table:comparison}
		\begin{minipage}[b]{0.45\linewidth}\centering
			\begin{tabular}{|l|l|l|}
				\hline
				Method  & $D$ (\%) & $VOE$ (\%) \\ \hline
				Christ et al.~\cite{Christ2016} & 94.3 & 10.7   \\ \hline
				Erdt et al.~\cite{Erdt2010}  & 94.6 (*) & 10.3   \\ \hline
				Li et al.~\cite{Li2013}  & 94.5 & 10.4 (*)   \\ \hline
				Li et al.~\cite{Li2015}  & 95.2 (*) & 9.15   \\ \hline
				Lu et al.~\cite{Lu2017}  & 95.0 (*) & 9.36   \\ \hline
				Sensor3D (full volume) & 95.4 & 8.79 \\ \hline
				Sensor3D (liver area)  & 95.9 & 7.87  \\ \hline
			\end{tabular}
		\end{minipage}
		\hspace{0.5cm}
		\begin{minipage}[b]{0.45\linewidth}
			\centering
			\begin{tabular}{|l|l|l|}
				\hline
				Method  & $D$ (\%) & $VOE$ (\%) \\ \hline
				Castro-Mateos et al.~\cite{Castro2015}  & 88.0 & 21.4 (*)  \\ \hline
				Forsberg et al.~\cite{Forsberg2015}  & 94.0 &  11.3 (*)  \\ \hline
				Hammernik et al.~\cite{Hammernik2015}  & 93.0 & 13.1 (*)  \\ \hline
				Korez et al.~\cite{Korez2015}  & 93.0 & 13.1 (*)  \\ \hline
				Seitel et al.~\cite{Seitel2015}  & 83.0 & 29.1 (*)  \\ \hline
				Sensor3D (full volume)  & 93.1 & 12.9 \\ \hline
				Sensor3D (vertebrae area **)  & 94.9 & 9.7 \\ \hline
			\end{tabular}
		\end{minipage}
	\end{table*}

	\subsection{Evaluations with different inter-slice distances}
	
	Table~\ref{table:two_fold_results} depicts the average Dice and volume overlap error scores for two folds of liver segmentation at different inter-slice distances $\mathbf{d}$. As expected, some irrelevant structures were partially segmented outside of the liver in a few cases thus lowering the scores when the full stack of volume slices is being considered. 
	
	The achieved results demonstrate that considering higher inter-slice distances is needed in order to get better segmentation performance. The lower scores for the 3 $mm$ inter-slice distance are caused by some scans in both the training and testing data where slice thicknesses exceed 3 $mm$. In such scans the extracted sequences may contain direct-consecutive slices therefore adding disturbance in the training by giving the network a wrong impression that the elements in the sequences are not really different. Thus, hindering the network to learn the inter-slice context for those training sequences properly. 
	

	\begin{figure*}[th!]
		\centering
		\includegraphics[width=0.45\textheight]{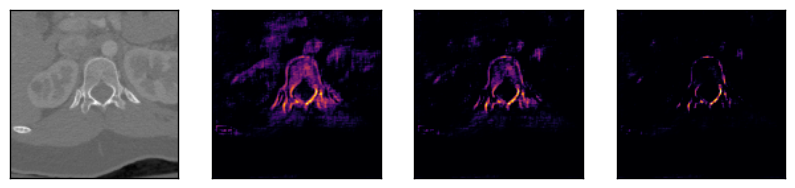}
		\includegraphics[width=0.45\textheight]{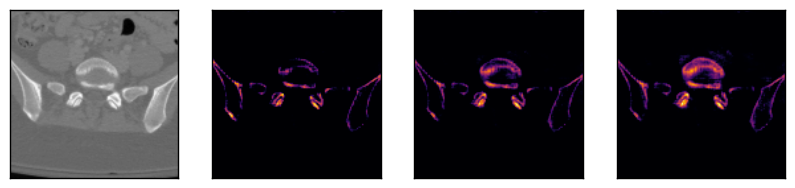}
		\caption{Examples of features extracted after the penultimate upsampling step (after $up_3$ layer in Table~\ref{table:architecture_details}) for two sample contexts containing the same~repeated~slice} 
		\label{fig:features}
	\end{figure*}
	
	We additionally analysed how the segmentation results of the models with sequences generated at various distances (shown in Table~\ref{table:two_fold_results}) differ in terms of statistical significance test scores. We performed pair-wise significance analysis using Wilcoxon signed-rank test for Dice scores on the test set. The results are shown in Table~\ref{table:significance} where the entries with values less than 0.01 correspond to pairs of models demonstrating statistically different significance in segmentation performance. Thus, the numbers complement and confirm the detailed results provided in Table~\ref{table:two_fold_results}: considering sequences of slices at the distances larger than $3 \,mm$ improves performance for the liver segmentation task significantly.  
		
	Some segmentation results at different vertebrae (top) and liver (bottom) areas are depicted in Fig.~\ref{fig:segmentations}. The red contour corresponds to the outline of the prediction, green to the ground-truth and yellow to the overlap of the outlines. 
	
	\subsection{Evaluations on the influence of the network capacity}
	
	Table~\ref{table:capacity_results} shows the detailed segmentation results of two-fold evaluations for architectures with different numbers of features in the convolutional layers and the \acrshort{CLSTM} units. The original configuration corresponds to the architecture shown in Table~\ref{table:architecture_details}. Two, four and eight times smaller configurations correspond to the architectures where the number of feature maps in the convolutional layers and \acrshort{CLSTM} blocks is two, four or eight times less than in the original configuration. 
	
	Performance results demonstrate that reducing capacity of the network by two times slightly worsens the results, however, the number of parameters in this configuration is almost four times less so that training time can be reduced significantly. Making the configuration even smaller inevitably worsens results especially in a more challenging Fold 1 where the test data consists of only a few challenging cases with multiple tumours inside or/and in a close proximity of the liver area.  
	
	\subsection{Performance of variants of the Sensor3D network}
	
	\subsubsection{2D modifications}
	
	In order to demonstrate that our Sensor3D network improves over similar 2D variants, we built and evaluated two additional architectures under the same training conditions on both folds in the liver segmentation task. 
	\textit{In the first} architecture, we set $\mathbf{o}=1$, thus, changed the input in a way that the network is fed with a sequence of single slices without context. \textit{In the second} architecture we did not change the slab size but removed the first \acrshort{CLSTM} and replaced the second one by the aggregation layer which would sum the incoming features along the time channel. Both architectures achieved similar average Dice scores of 84.3 \% and 85.6 \% (computed over two folds) when considering the organ area only. For the full scan scores of 73.1 \% and 74.5\% were achieved which are similar to the results of the U-Net performance reported by Christ et al.~\cite{Christ2016}. These scores are notably lower than the results demonstrated by Sensor3D. It shows that learning 3D context is crucial for achieving a better performance.

	\subsubsection{Unidirectional modification} 
	We built and evaluated a unidirectional modification of the Sensor3D architecture under the same training conditions on both folds in the liver segmentation task. In this architecture we have replaced $bidir_1$ and $bidir_2$ layers in Table~\ref{table:architecture_details}  with unidirectional \acrshort{CLSTM} blocks. The model achieved the Dice score of 93.53 \% when considering the organ area only and 91.50 \% in the full volume. The achieved scores are significantly lower than the ones reached by both the state-of-the-art methods and Sensor3D on the same task in particular which hardens the assumption that bidirectional modification is beneficial for this architecture.

	\subsection{Comparison with state-of-the-art methods}

	Table~\ref{table:comparison} (left) compares our approach with the state-of-the-art methods trained and tested on the same 3Dircadb dataset. Though our model is trained only on the parts of the volumes where the liver is present (from 33\% to 95\% of slices in different scans, and in 71\% of slices on average across all scans), it can still reach competitive and in many cases better results when evaluated against 2D and 3D approaches considering both the liver area and the full volume. 
	
	To demonstrate that our method generalizes on other organs as well, we have trained and evaluated the network on the CSI 2014 dataset on the vertebrae segmentation task. Table~\ref{table:comparison} (right) compares performance of our approach with several state-of-the-art methods. It is worth noting that some vertebrae which are not present in the ground-truth annotations are still segmented by our network thus causing lower scores in the cases when the full volume is considered.	

	\subsection{Visual feature inspection}
	
	In order to visually demonstrate the sequential nature of the features learnt by our model, we performed the following test. We passed two sequences to the network (both for vertebrae), each containing three identical slices (first column in Fig.~\ref{fig:features}). The columns show some of the features extracted after the penultimate upsampling step (after $up_3$ layer in Table~\ref{table:architecture_details}) before passing them to the final bidirectional \acrshort{CLSTM} block. The visualization shows that the layers respond differently to the same input element, activating different parts of the organ of interest. The brighter colour intensities correspond to higher activations. Comparing the rows, it shows that the network is able to learn spatial correlations in both directions.

	\section{Conclusions}
	In this paper we proposed Sensor3D, a general, robust, end-to-end U-Net-like hybrid architecture combining time-distributed convolutions, pooling, upsampling layers and bidirectional \acrshort{CLSTM} blocks. To demonstrate generalization of our approach, we evaluated the model on liver and additionally vertebrae segmentation task on the publicly available 3Dircadb and CSI 2014 datasets. Quantitative evaluations of the 2D variants of the Sensor3D network, statistical significance test, evaluation on the network capacity indicate that the \acrshort{CLSTM} boosts overall performance. Visual inspection of the model activation on the sequences containing the same repeated slices shows firing of different areas in the organs therefore empirically proving the sequential nature of the learnt features.  Contrary to the state-of-the-art models, our network does not require full input volumes for neither training nor inference. Our network shows competitive and often superior performance on the considered liver and vertebrae segmentation tasks despite that it was trained only on slabs of the training volumes. For future work, we plan to apply our algorithm to other imaging modalities and organs in a multi-task manner.  
	
	%
	%
	\bibliographystyle{unsrt}
	\bibliography{references}{}
	%
	
	
	
	
	
	

\end{document}